\documentclass[fleqn,10pt]{wlscirep}
\usepackage{multirow}
\usepackage[super,sort&compress,comma]{natbib}
\title{Medical Diagnosis From Laboratory Tests by Combining Generative and Discriminative Learning}

\author[1]{Shiyue Zhang}
\author[2]{Pengtao Xie}
\author[3]{Dong Wang}
\author[4,*]{Eric P. Xing}

\affil[1]{Beijing University of Posts and Telecommunications, Beijing, China}
\affil[2]{Machine Learning Department, Carnegie Mellon University, Pittsburgh, USA}
\affil[3]{Center for Speech and Language Technologies, Tsinghua University, Beijing, China}
\affil[4]{Petuum Inc, Pittsburgh, USA}
\affil[*]{eric.xing@petuum.com}

\begin{abstract}
A primary goal of computational phenotype research is to conduct medical diagnosis. In hospital, physicians rely on massive clinical data to make diagnosis decisions, among which laboratory tests are one of the most important resources. However, the longitudinal and incomplete nature of laboratory test data casts a significant challenge on its interpretation and usage, which may result in harmful decisions by both human physicians and automatic diagnosis systems. In this work, we take advantage of deep generative models to deal with the complex laboratory tests. Specifically, we propose an end-to-end architecture that involves a deep generative variational recurrent neural networks (VRNN) to learn robust and generalizable features, and a discriminative neural network (NN) model to learn diagnosis decision making, and the two models are trained jointly. Our experiments are conducted on a dataset involving 46,252 patients, and the 50 most frequent tests are used to predict the 50 most common diagnoses. The results show that our model, VRNN+NN, significantly ($p<0.001$) outperforms other baseline models. Moreover, we demonstrate that the representations learned by the joint training are more informative than those learned by pure generative models. Finally, we find that our model offers a surprisingly good imputation for missing values.

\end{abstract}
\begin{document}
\flushbottom
\maketitle
%
%
\thispagestyle{empty}

\section*{Introduction}

Misdiagnosis, where a diagnostic decision is made inaccurately, widely occurs. There are approximately 12 million US adults experiencing diagnosis errors every year, half of which could be harmful \cite{Singhbmjqs-2013-002627}. As many as 40,500 adult patients in an intensive care unit (ICU) die with an misdiagnosis annually \cite{winters2012diagnostic}. A major source of misdiagnosis is sub-optimal interpretation and usage of clinical data \cite{schiff2009diagnostic, DiagError, Singh2013Types, doi:10.7326/0003-4819-145-7-200610030-00006}. Nowadays, physicians are overwhelmed by a large amount of medical data including laboratory tests, vital signs, clinical notes, medication prescriptions, etc. Among all kinds of clinical data, laboratory tests play an important role. According to American Clinical Laboratory Association, laboratory tests guide more than 70\% of the diagnostic decisions. Unfortunately, comprehensively understanding the laboratory test results and discovering the underlying clinical implications are not easy. Incorrect interpretation of laboratory tests is a major breakdown point
in the diagnostic process \cite{doi:10.7326/0003-4819-145-7-200610030-00006, DiagError, schiff2009diagnostic, Singh2013Types}.

The reasons why laboratory tests are difficult to understand are two-fold. First, missing values are pervasive. It is typical that at a certain time point, only a small subset of laboratory tests are examined, leaving the values of most tests missed. These data missing prevents physicians from getting a full picture of patients' clinical states, leading to sub-optimal decisions. Second, the laboratory test values have a complex multivariate time-series structure: during an in-hospital stay, multiple laboratory tests are examined at a particular time, and the same test may be examined multiple times at different time points. These multivariate temporal data exhibits complicated patterns along the dimensions of both time and tests. Learning these patterns is highly valuable for diagnosis, but it is technically challenging.

In this work, we study how to leverage the ability of machine learning (ML) in automatically distilling patterns from complex, noisy, incomplete and irregular laboratory test data to address the above-mentioned issues, and build an end-to-end diagnostic model to assist with diagnosis decisions. Previous studies have applied ML to perform diagnosis based on laboratory tests. In these approaches, three major tasks – handling missing values, discovering patterns from multivariate time-series and predicting diseases – were often conducted separately. However, these three tasks are tightly coupled and can mutually benefit each other. On one hand, better imputation of missing values leads to the discovery of more informative patterns, which boosts the accuracy of diagnosis. On the other hand, during model training, the supervision of diagnosis provides a guidance of pattern discovery, which further influences the imputation of missing values, tailoring the discovered patterns and imputed values to be suitable for a diagnosis task. Performing these tasks separately fails to consider their synergistic relationships and hence leads to sub-optimal solution. Another limitation exists in previous studies is that they often proposed in a discriminative structure that cannot well address the missing value problem and learn generalizable patterns in principle.

\section*{Contributions}

In this paper, we develop an end-to-end deep neural model to perform diagnosis based on laboratory tests. Our model seamlessly integrates three tasks together: imputing missing values, discovering patterns from multivariate time-series data and predicting diseases, and perform them jointly. Our model combines two major learning paradigms in machine learning: generative learning and discriminative learning, where the generative learning component is utilized to deal with missing values and discover robust and generalizable patterns and the discriminative learning component is used for predicting diseases based on the patterns discovered in generative learning. We evaluate the proposed model on 46,252 patient visits in the ICU and demonstrate that our model achieves (1) significantly ($p<0.001$) better diagnosis performance than baseline models, (2) better imputation of missing values, and (3) better discovery of patterns from the laboratory test data.

\section*{Related Works}

Lasko \emph{et al.} proposed to use a Gaussian process to model the longitudinal Electronic Medical Records (EHR) and used a standard auto-encoder (AE) to learn hidden features from raw inputs \cite{lasko2013computational}, and Ghassemi \emph{et al.} further introduced a multi-task Gaussian process to model the clinical data \cite{ghassemi2015multivariate}. Che \emph{et al.} and Miotto \emph{et al.} took advantage of denoising auto-encoder (DAE) \cite{vincent2008extracting} to learn hidden representations and then used these representations to make diagnose \cite{che2015deep, miotto2016deep}. In recent years, recurrent neural network (RNN) has demonstrated its superiority in modeling longitudinal data, like natural language \cite{mikolov2010recurrent} and speech signals \cite{graves2013speech}. RNN has also been utilized for medical diagnosis. Lipton \emph{et al.} proposed a medical diagnosis model based on long short-term memory (LSTM) \cite{hochreiter1997long} recurrent neural networks and obtained performance better than some strong baselines \cite{lipton2015learning, lipton2015phenotyping}. In this work, missing values were addressed by heuristic forward- and back-filling. Their later work continued the LSTM-based diagnosis, but used a missing value indicator as a part of the inputs. They found missing value patterns can help with diagnosis performance \cite{lipton2016modeling}. Choi \emph{et al.} proposed a 'Doctor AI' that used previous diagnoses to predict future diagnosis \cite{choi2016doctor}. In their another work, a neural attention mechanism was further introduced \cite{choi2016retain}. Che \emph{et al.} proposed a diagnosis model based on stacked DAE and LSTM, and Gradient Boosting Trees were introduced to make the learned features more interpretable \cite{che2015distilling}. Their later work modified the structure of Gated Recurrent Units (GRU) to deal with incomplete inputs \cite{che2016recurrent}.

All the above work are based on discriminative models, which cannot address the missing value problem very well in principle. A generative model usually works better in this aspect. A typical generative model is the Gaussian Mixture Model (GMM), with which missing values can be easily addressed by the Expectation-Maximum (EM) algorithm \cite{Ghahramani1994Learning, ghahramani1994supervised}. For instance, Marlin \emph{et al.} employed GMMs to discovery patterns from clinical data and conducted mortality prediction \cite{Marlin2012Unsupervised}. These generative models, however, are mostly shallow, linear and Gaussian.
Recently, researchers proposed several deep generative models, e.g., variational auto-encoder (VAE) \cite{kingma2013auto} and variational RNN (VRNN) \cite{chung2015recurrent}. Compared to conventional generative models, VAE and VRNN can model more complex conditional distributions, hence representing more complex patterns. Our work will utilize these deep generative models. However, generative models are not task-oriented \cite{yakhnenko2005discriminatively, bernardo2007generative, yogatama2017generative}. Therefore, we propose an end-to-end approach that combines the advantages of both types of models and train them in a joint fashion. This can be seen as either a generative learning with the discriminative target as a guidance, or a discriminative learning with the generative target as a regularization.

\section*{Methods}
\subsection*{Data Preparation}
\begin{figure}[t]
	\centering
	\includegraphics[width=0.65\linewidth]{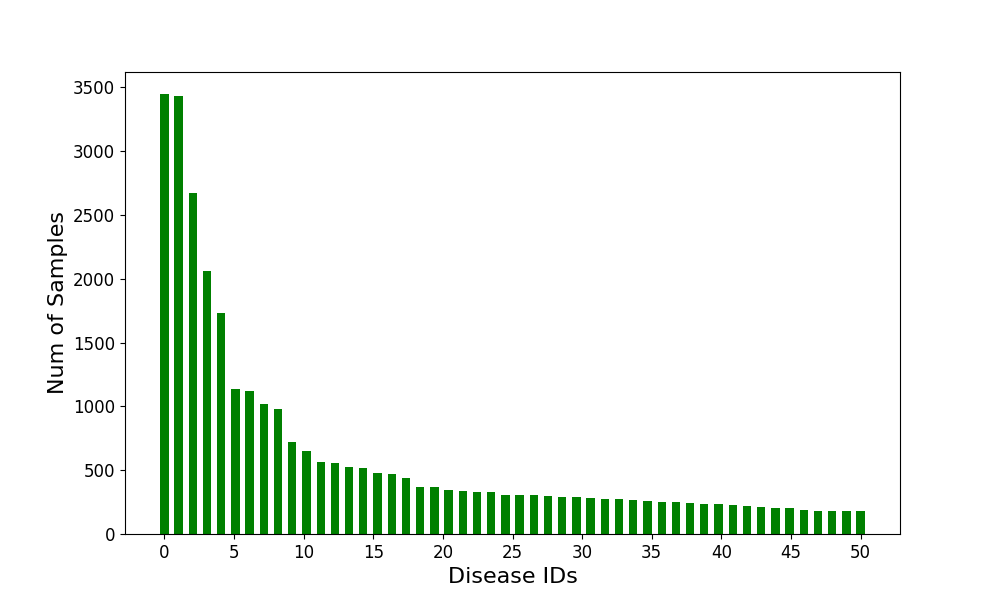}
	\caption{Number of Samples over Disease IDs}
	\label{fig1}
\end{figure}

The data used in this study, opened by MIMIC-III\cite{johnson2016mimic}, is publicly available. It was derived from the laboratory tests of 46,252 patients. It contains both in-hospital and outpatient records. Each in-hospital episode has 1 to 39 corresponding ICD-9 (International Classification of Diseases, 9th Edition) codes, and in this study, only the primary diagnoses will be considered. It amounts to 2,789 different diagnoses and 513 unique laboratory tests. As some diagnoses and tests are quite rare, we limit our study to the 50 most frequent diagnoses and the 50 most frequent laboratory tests. We group the test results by day, and finally, we get 30,931 temporal sequences of in-hospital records, and each of them is labeled by a disease ID, from 0 to 49. The lengths of the temporal sequences range from 2 to 171, and we focus on the latest 100 days. Figure~\ref{fig1} shows the number of samples over the 50 disease IDs. We random split the dataset for 5 times, and each time we keep the proportion of training (Train), development (Dev), testing (Test) sets as 65\%:15\%:20\%. Hence, the numbers of samples in these three sets are 20,105, 4,640, 6,186, respectively.

\begin{figure}[t]
	\centering
	\includegraphics[width=0.65\linewidth]{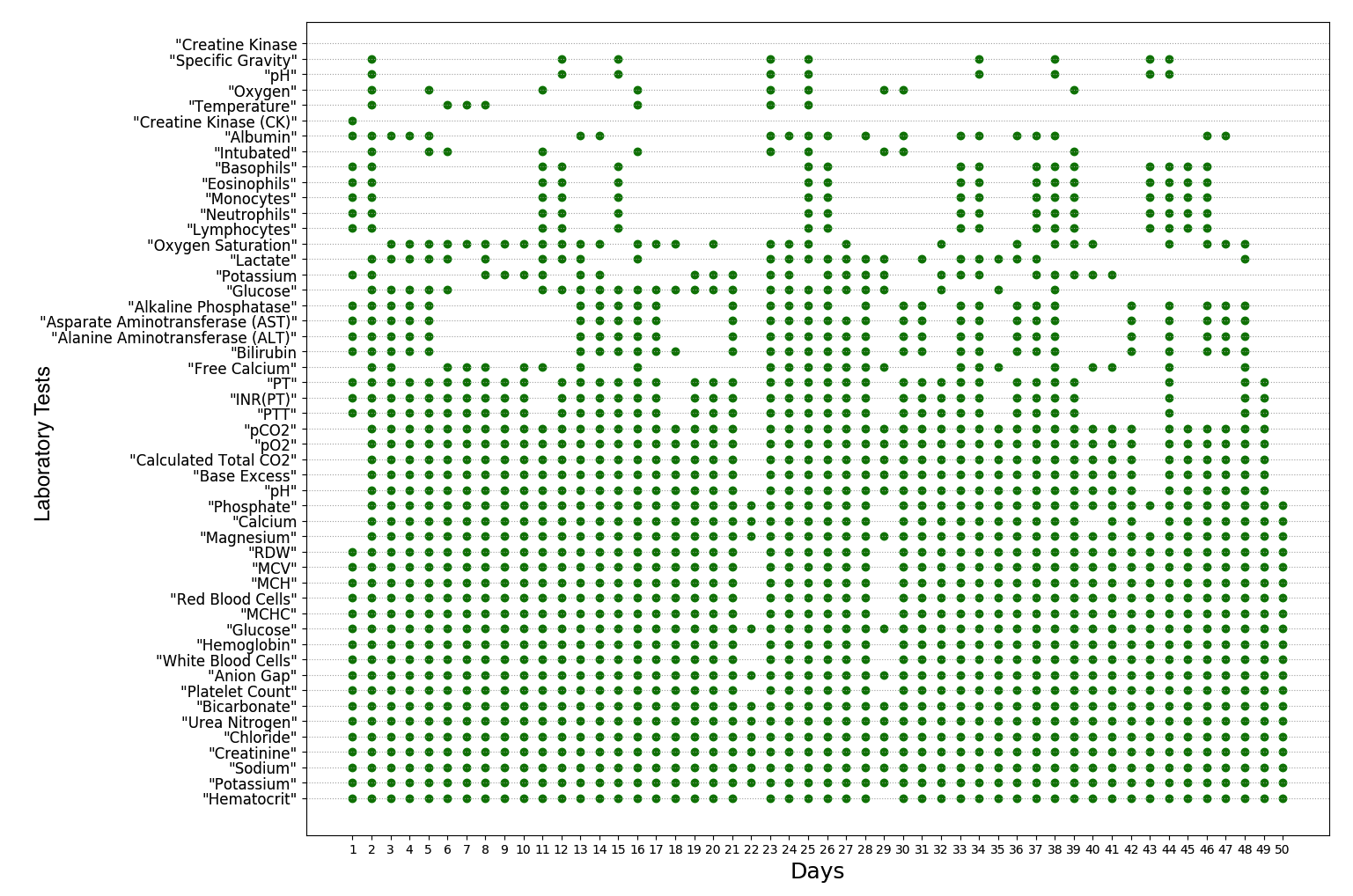}
	\caption{An example of a patient's laboratory test records. The y axis corresponds to 50 laboratory tests we used. The x axis denotes the time of the records. The green dot means there is a value, otherwise it is missing.}
	\label{fig2}
\end{figure}

Some of the tests are valued by discrete categories, like "ABNORMAL" and "NORMAL". We change these categories to integers, like 0 for "ABNORMAL", 1 for "NORMAL". Test results are normalized by the Z-normalization, i.e. values of each test are subtract by the mean and divided by the standard deviation. Note that a patient cannot do every test every day, so missing values are pervasive in our data. In Figure~\ref{fig2}, we present an example of a patient's laboratory test records. It can be seen that there are a lot of missing values. A simple statistics show that in the whole dataset, the average missing value rate is about 54\%, i.e. on average, only 27 of the 50 laboratory tests have values in a patient's one-day record. In our experiments, initially we impute the missing values with 0. After applying Z-normalization, the mean of values change to zero. So, zero imputation equals to mean imputation. Moreover, since our models are situated in neural network framework, zero inputs will not introduce additional bias in computation. Note that in baseline models, this zero-imputation behaves as the solution of missing value problem, while in our models, it behaves as indicators of the missing values, and missing values will be further addressed by deep generative models.

\subsection*{Model Architectures}
Given N i.i.d data $\textbf{X} = \{\textbf{x}_n\}_{n=1}^N$, each $\textbf{x}_n$ is a temporal sequence of $T$ in length, i.e. $\textbf{x}_n = \{x_1, x_2, ..., x_T\}$. Any $t\in[1,T]$, $x_t \in\mathbb{R}^M$, where $M$ is the dimension of input features. Meanwhile, there is a class label $y_n$ for each $\textbf{x}_n$, and our purpose is to predict the class labels accurately. Specifically, $\textbf{x}_n$ is the longitudinal laboratory test records of a patient, and each $x_t$ is a one-day record. $y_n$ is the primary diagnosis. We propose two models in this study, denoted by VAE+NN and VRNN+NN, the former a static model that can demonstrate the contribution of deep generative models, while the latter a temporal model that extends the deep generative learning approach to learn long-term temporal dependency.

\subsubsection*{VAE+NN}

In this model, we address the temporal records by simply averaging the vectors at all time points, i.e. $\tilde{x}_n = average(x_1, x_2, ..., x_T)$. Although the averaging operation alleviates the missing value problem to some extent, the rate of the missing values in $\tilde{x}_n$ is still high (about 29\%). To further deal with the missing values and capture the complex patterns in data, we propose a VAE+NN model, where a VAE is the generative model used to handle missing values and discover patterns, and a standard neural network (NN) is used as a classifier, as shown in Figure~\ref{fig3} (a).

Referring to the idea of VAE \cite{kingma2013auto}, suppose that $\tilde{x}_n$ is generated from a latent variable $z_n$, then the joint probability is defined as:
\begin{equation}
p(X, Z) = \prod\limits_{n=1}^Np(z_n)p(\tilde{x}_n|z_n).
\end{equation}
Let the prior over $z_n$ is a centered isotropic multivariate Gaussian distribution, i.e. $p(z_n) = \mathcal{N}(z;\textbf{0},\textbf{I})$. Further assume that with a fixed $z_n$, $\tilde{x}_n$ also follows a Gaussian distribution, and suppose different dimensions of $\tilde{x}_n$ are independent, then the generation of $\tilde{x}_n$ is defined as:
\begin{equation}
p(\tilde{x}_n|z_n) = \mathcal{N}(\mu_x, diag(\sigma^2_x)), where [\mu_x, \sigma_x] = \varphi_{\tau}^{dec}(z_n),
\end{equation}
where $\mu_x$ and $\sigma_x$ are parameters of the conditional distribution. $\varphi_{\tau}^{dec}$ is a feed-forward neural network. Then, the expectation of posterior distribution $p(z_n|\tilde{x}_n)$ is fed into a discriminative network $\varphi_d$ to perform classification, formulated as follows:
\begin{equation}
p(\hat{y}_n|\textbf{x}_n) = softmax(\varphi_d(E(z_n|\tilde{x}_n))).
\end{equation}
where $softmax$ is the softmax function, written by $\frac{e^k} {\sum_{k}e^k}$.

\subsubsection*{VRNN+NN}

\begin{figure}[t]
	\centering
	\includegraphics[width=0.75\textwidth]{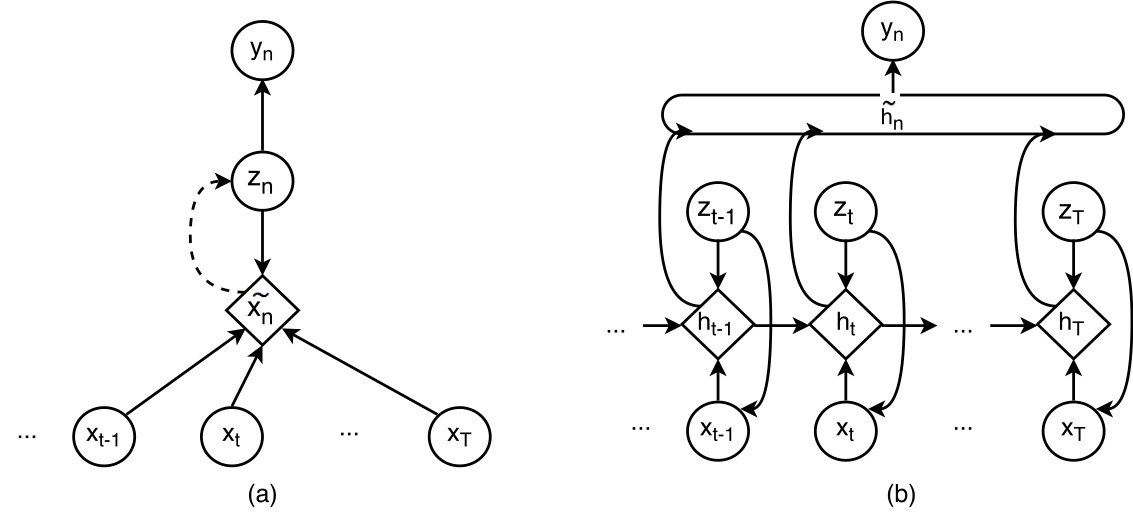}
	\caption{Model Architectures of Our Models: (a) is the architecture of VAE+NN, and (b) is the architecture of VRNN+NN.}
	\label{fig3}
\end{figure}

To better capture the long-term dependency of clinical records, previous work introduced RNN into this task \cite{lipton2015learning, lipton2016modeling}. However, the internal transition structure of RNN is entirely deterministic which cannot efficiently model the complexity in highly structured sequential data. Therefore, Chung \emph{et al.} extended the VAE idea to a recurrent framework for modeling high-dimensional sequences, leading to a variational RNN (VRNN)\cite{chung2015recurrent}. We propose a VRNN+NN model as shown in Figure~\ref{fig3} (b), which involves a VRNN to generate sequential hidden features and an NN model to make decisions based on the average of these hidden features.

Assuming that the generation of each $x_t$ in $\textbf{x}_n$ is conditioned on a latent variable $z_t$ and the previous state $h_{t-1}$, the joint probability is defined as follows.
\begin{equation}
p(X, Z) = \prod\limits_{n=1}^N\prod\limits_{t=1}^Tp(z_t; h_{t-1})p(x_t|z_t; h_{t-1}),
\end{equation}
where $p(z_t; h_{t-1})$ is the prior of the latent variable and $p(x_t|z_t; h_{t-1})$ is the conditional probability of the observations. We assume that the prior distribution $p(z_t; h_{t-1})$ follows a Gaussian distribution, and different dimensions of $z_t$ are independent, i.e. the covariance matrix is diagonal. Therefore, the prior distribution is defined as follows.
\begin{equation}
p(z_t; h_{t-1}) = \mathcal{N}(\mu_{0, t},diag(\sigma^2_{0, t})), where [\mu_{0, t}, \sigma_{0, t}] = \varphi_{\tau}^{prior}(h_{t-1}),
\end{equation}
where $\mu_{0, t}$ and $\sigma_{0, t}$ represents the mean and the standard variance of the prior distribution, respectively. Again, assume different dimensions of $x_t$ are independent, then the conditional probability for the observation is written by:
\begin{equation}
p(x_t|z_t; h_{t-1}) = \mathcal{N}(\mu_{x, t}, diag(\sigma^2_{x, t})), where [\mu_{x, t}, \sigma_{x, t}] = \varphi_{\tau}^{dec}(z_t, h_{t-1}),
\end{equation}
where $\mu_{x, t}$ and $\sigma_{x, t}$ are parameters of the generating distribution. $\varphi_{\tau}^{prior}$ and $\varphi_{\tau}^{dec}$ are neural networks. The hidden states are updated by LSTM \cite{hochreiter1997long} cells denoted by $f_\theta$, so the recurrence equation is defined as:
\begin{equation}
h_t = f_\theta(x_t, z_t, h_{t-1}).
\end{equation}
We take the average of the hidden states, $\tilde{h}_n = average(h_1, h_2, ..., h_T)$, as the input of a discriminative NN $\varphi_{d}$.
\begin{equation}
p(\hat{y}_n|\textbf{x}_n) = softmax(\varphi_{d}(\tilde{h}_n))
\end{equation}

\subsection*{Inference \& Learning}

\subsubsection*{VAE+NN}
The involving of neural networks makes the true posterior $p(z_n|\tilde{x}_n)$ is intractable. Resort to variational inference methods, we use $q(z_n|\tilde{x}_n)$ to approximate the true posterior which is defined in the fashion of:
\begin{equation}
q(z_n|\tilde{x}_n) = \mathcal{N}(\mu_z, diag(\sigma^2_z)), where [\mu_z, \sigma_z] =
\varphi_{\tau}^{enc}(\tilde{x}_n),
\end{equation}
where $\varphi_{\tau}^{enc}$ is a feed-forward neural network. Denote the parameters involved in VAE as $\theta_{g}$, learning target is $\hat{\theta}_{g} = argmax_{\theta_g}(\sum_{n}logp(\tilde{x}_n|\theta_{g}))$ which can be redefined as the lower bound of $logp(\tilde{x}_n|\theta_{g})$:
\begin{equation}
\mathcal{L}_{g}(\theta_{g}; X) = \sum^N_{n}(-KL(q(z_n|\tilde{x}_n)||p(z_n)) +  E_{q(z_n|\tilde{x}_n)}[logp(\tilde{x}_n|z_n)]).
\end{equation}

Since we assume different dimensions of $\tilde{x}$ are independent, $p(\tilde{x}_n|z_n) = \prod\limits_{m} p(\tilde{x}_{n,m}|z_n)$. Missing items in $\tilde{x}_n$ should be dropped in the probability, i.e. $p(\tilde{x}_n|z_n) = \prod\limits_{m \notin \tilde{\textbf{m}}} p(\tilde{x}_{n,m}|z_n)$, $\tilde{\textbf{m}}$ is the indexes of missing items.  So, the generative loss can be rewritten as follows:
\begin{equation}
\mathcal{L}_{g}(\theta_{g}; X) = \sum^N_{n}(-KL(q(z_n|\tilde{x}_n)||p(z_n)) +  E_{q(z_n|\tilde{x}_n)}[\sum_{m \notin \tilde{\textbf{m}}}logp(\tilde{x}_{n, m}|z_n)]).
\end{equation}
Also, there is a discriminative loss from classification, which is the cross entropy between the true disease ID $y_n$ and model predicted posterior probability. $\theta_{d}$ denotes the parameters in discriminative network.
\begin{equation}
\mathcal{L}_{d}(\theta_{d}; X, Y) = \sum^N_n log p(y_n|\textbf{x}_n).
\end{equation}
Therefore, the overall loss function of the VAE+NN model is given by the sum of the two costs, denote $\theta = \{\theta_{g}, \theta_{d}\}$.
\begin{equation}
\mathcal{L}(\theta; X, Y) = \mathcal{L}_{d}(\theta_{d}; X, Y) + \eta * \mathcal{L}_{g}(\theta_{g}; X),
\end{equation}
where $\eta$ is a trade-off parameter of generative loss, because our final target is classification, and thus the generative loss can be taken as a regularization term.

\subsubsection*{VRNN+NN}
Similarly, the true posterior $p(z_t|x_t)$ is intractable, $q(z_t|x_t)$ is used to approximate it. $q(z_t|x_t)$ and generative loss $\mathcal{L}_{g}(\theta_{g}; X)$ are defined as follows:
\begin{equation}
q(z_t|x_t) = \mathcal{N}(\mu_{z, t}, diag(\sigma^2_{z, t})), where [\mu_{z, t}, \sigma_{z, t}] =
\varphi_{\tau}^{enc}(x_t, h_{t-1}),
\end{equation}
\begin{equation}
\mathcal{L}_{g}(\theta_{g}; X) = \sum^N_{n}\sum^T_{t}(-KL(q(z_t|x_t)||p(z_t)) +  E_{q(z_t|x_t)}[logp(x_t|z_t)]).
\end{equation}
Also, since different dimensions of $x_t$ are independent, $p(x_t|z_n) = \prod\limits_{m} p(x_{t,m}|z_t)$. Denote the indexes of missing values in $x_t$ is $\tilde{\textbf{m}}_t$, then:
\begin{equation}
\mathcal{L}_{g}(\theta_{g}; X) = \sum^N_{n}\sum^T_{t}(-KL(q(z_t|x_t)||p(z_t)) +  E_{q(z_t|x_t)}[\sum_{m \notin \tilde{\textbf{m}}_t}logp(x_{t, m}|z_t)]),
\end{equation}
The overall loss function is also defined as the combination of generative loss and discriminative loss.
\begin{equation}
\mathcal{L}_{d}(\theta_{d}; X, Y) = \sum^N_n log p(y_n|\textbf{x}_n),
\end{equation}
\begin{equation}
\mathcal{L}(\theta; X, Y) = \mathcal{L}_{d}(\theta_{d}; X, Y) + \eta * \mathcal{L}_{g}(\theta_{g}; X).
\end{equation}

\subsection*{Baselines}
In this section, we will build several baseline models for the comparative study, denoted by NN, AE+NN, RNN+NN, respectively. NN and AE+NN models are used to compare with the VAE+NN model, we'd like to see if deep generative models can have a better performance when representing single feature vectors. RNN+NN model is similar as the model structure in previous studies \cite{lipton2015learning,lipton2016modeling}, and it is used to compare with our VRNN+NN model.

\subsubsection*{NN}

As shown in Figure~\ref{fig4} (a), 'NN' model is a simple multi-layer perceptron (MLP). The model and the learning target are defined as follows:
\begin{equation}
p(\hat{y}_n|\textbf{x}_n) = softmax(\varphi_{d}(\tilde{x}_n)),
\end{equation}
\begin{equation}
\mathcal{L}(\theta; X, Y) = \sum^N_n log p(y_n|\textbf{x}_n).
\end{equation}

\subsubsection*{AE+NN}
\begin{figure}[t]
	\centering
	\includegraphics[width=1.0\textwidth]{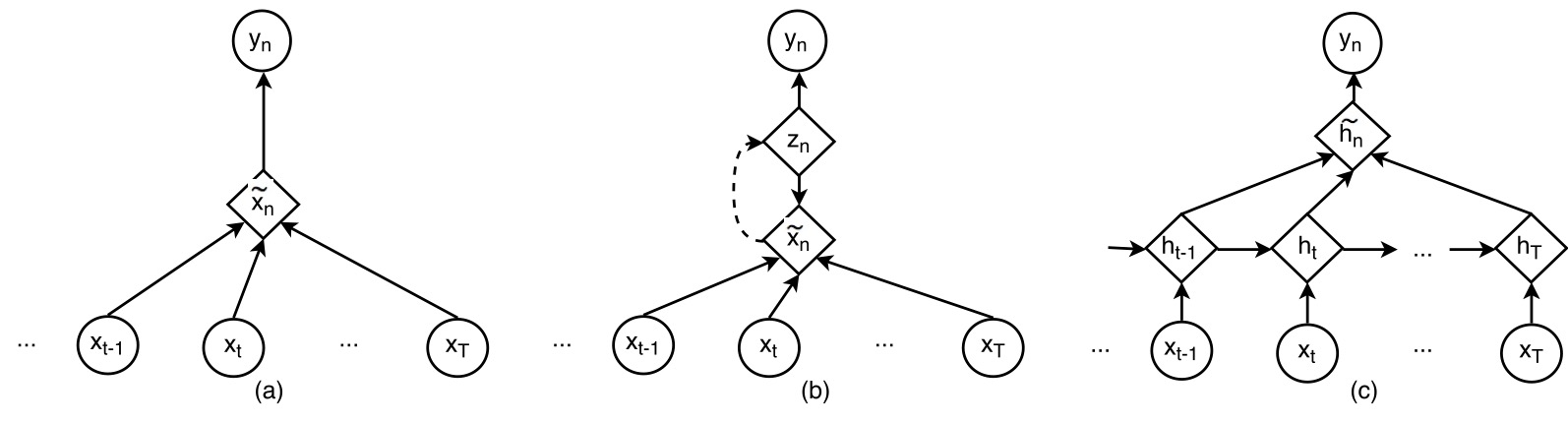}
	\caption{Model Architectures of Baseline Models: (a) is the architecture of NN, (b) is the architecture of AE+NN, and (c) is the architecture of RNN+NN.}
	\label{fig4}
\end{figure}

To demonstrate the superiority of VAE, we present an AE+NN baseline that is based on the standard auto-encoder, shown in Figure~\ref{fig4} (b). AE is similar to VAE, but its structure is deterministic, so it is less generative. In this model, we also combine the loss from AE and NN. We use mean squared error (MSE) as the training objective for AE. Also, missing items are not included in the loss function. In summary, the model and the learning target are defined as follows:
\begin{equation}
z_n = \varphi_{\tau}^{enc}(\tilde{x_n}),
\end{equation}
\begin{equation}
\hat{x}_n = \varphi_{\tau}^{dec}(z_n),
\end{equation}
\begin{equation}
p(\hat{y}_n|\textbf{x}_n) = softmax(\varphi_{d}(z_n)),
\end{equation}
\begin{equation}
\mathcal{L}(\theta; X, Y) = \sum^N_n (log p(y_n|\textbf{x}_n) + \eta * \sum_{m \notin \tilde{\textbf{m}}}(\tilde{x}_{n, m} - \hat{x}_{n,m})^2).
\end{equation}

\subsubsection*{RNN+NN}
The RNN+NN model is shown in Figure~\ref{fig4} (c). In this model, an RNN processes the raw temporal features, and the average of the hidden state $\tilde{h}_n$ is used as the input of the NN. $f_\theta$ also denotes the recurrent computation of the LSTM cells. The model can be formulated as follows:
\begin{equation}
h_t = f_\theta(x_t, h_{t-1}),
\end{equation}
\begin{equation}
p(\hat{y}_n|\textbf{x}_n) = softmax(\varphi_{d}(\tilde{h}_n)),
\end{equation}
\begin{equation}
\mathcal{L}(\theta; X, Y) = \sum^N_n log p(y_n|\textbf{x}_n).
\end{equation}

\subsection*{Implement Details}
In our experiments, models are implemented on Tensorflow r1.0. All $\varphi_{\tau}$ and $\varphi_{d}$ are feed-forward neural networks with one hidden layer and ReLU activations. The size of hidden layer is set to 64. We use Adam\cite{kingma2014adam} as the optimizer, where the learning rate is 0.0005 and the learning rate decay is 0.99. The trade off parameter $\eta$ is set to 0.5 in all the experiments.

\subsection*{Evaluation Metrics}

\subsubsection*{F1 score}
Generally, F1 score is defined as $\frac{2*precision*recall}{precision+recall}$, where $precision$ is the number of correct positive results divided by the number of all positive results, and $recall$ is the number of correct positive results divided by the number of positive results that should have been returned. Since our task is a 50-class classification, we use three kinds of F1 scores: micro-F1, macro-F1, macro-F1-weighted (macro-F1-w). Micro-F1 is computed from flatten $\hat{Y}$ and $Y$. Macro-F1 is the arithmetic mean of F1 scores of different classes. Macro-F1-weighted is the weighted mean of F1 scores of different classes, and the weight of class $i$ is defined as $\frac{n_i}{N}$, where $n_i$ is the number of samples of class $i$ in testing set. F1 score ranges from 0 to 1, and the best performance achieves when F1 score equals 1. We simulate a blind prediction, i.e. equal probabilities on different classes, the results are micro-F1 is 0.111, macro-F1 is 0.004, macro-F1-weighted is 0.022.

\subsubsection*{AUC}
AUC is the area under the ROC curve. Similarly, we also use three AUCs in evaluation: micro-AUC, macro-AUC, macro-AUC-weighted (macro-AUC-w). AUC also ranges from 0 to 1, and the best performance achieves when AUC  equals 1. And in the blind prediction, all AUCs equal 0.5.

\section*{Results}
Table~\ref{tab1} shows the diagnosis performances of three sets of experiments. The top set of values are the performances of different models on the diagnosis task, measured by different variants of F1 values and AUCs. Additionally, to test if the joint training can result in better representations compared to unsupervised generative models, the representations derived from VAE, VAE+NN, VRNN and VRNN+NN are used to train a new NN model for diagnosis decision. Performances are presented in the middle set. Finally, in the bottom set, to compare our VRNN+NN model's ability in dealing with missing values with some heuristic imputation methods, we investigate four imputation methods here. "zero" is the default approach for baseline models, and "last\&next", "row mean" and "NOCB" are three best known imputation methods according to the previous study \cite{engels2003imputation}: "last\&next" is the average of the last known and next known values; "row mean" is the mean of patient's values before and after; "NOCB" is the next observation carried backward.

To evaluate if the performance in Table~\ref{tab1} is reliable, we apply paired t-test to check if the performance difference among different models is statistically significant. The results are shown in Table~\ref{tab2}.

Since deep generative models can reconstruct input data, we conjecture that our VRNN+NN model has the potential to impute missing values better. To test this conjecture, we first randomly drop 10\% of the values from the original data, and then use the trained  VRNN+NN to impute the intentionally dropped values. The results in terms of MSE are shown in Table~\ref{tab3}, where the MSE values of the heuristic imputation methods are also presented. The paired t-test results of these methods are shown as well.

\begin{table}[t]
	\centering
	\begin{tabular}{ccccccc}
		\multicolumn{7}{c}{Diagnosis performances of different models} \\
		\hline
		Model & Micro-F1 & Macro-F1 & Macro-F1-w & Micro-AUC & Macro-AUC & Macro-AUC-w\\
		\hline
		NN & 0.376 $\pm$ 0.004  & 0.221 $\pm$ 0.003 & 0.347 $\pm$ 0.005 & 0.939 $\pm$ 0.001 & 0.905 $\pm$ 0.001 & 0.913 $\pm$ 0.001\\
		AE+NN & 0.366 $\pm$ 0.004 & 0.219 $\pm$ 0.002 & 0.344 $\pm$ 0.002 & 0.938 $\pm$ 0.001 & 0.903 $\pm$ 0.002 & 0.912 $\pm$ 0.001\\
		VAE+NN & 0.374 $\pm$ 0.003 & 0.226 $\pm$ 0.005 & 0.352 $\pm$ 0.004 & 0.941 $\pm$ 0.000 & 0.908 $\pm$ 0.001 & 0.916 $\pm$ 0.001\\
		RNN+NN & 0.395 $\pm$ 0.004 & 0.248 $\pm$ 0.003 & 0.373 $\pm$ 0.003 & 0.945 $\pm$ 0.003 & 0.918 $\pm$ 0.004 & 0.923 $\pm$ 0.003\\
		VRNN+NN & \textbf{0.426 $\pm$ 0.002} & \textbf{0.291 $\pm$ 0.006} & \textbf{0.407 $\pm$ 0.002} & \textbf{0.958 $\pm$ 0.000} & \textbf{0.937 $\pm$ 0.000} & \textbf{0.938 $\pm$ 0.001}\\
		\hline
		\multicolumn{7}{c}{Performance of features derived from different models (with a simple NN classifier)} \\
		\hline
		E($z_n$)(VAE) & 0.363 $\pm$ 0.004  & 0.195 $\pm$ 0.004 & 0.326 $\pm$ 0.003 & 0.936 $\pm$ 0.001 & 0.896 $\pm$ 0.003 & 0.906 $\pm$ 0.002\\
		E($z_n$)(VAE+NN) & 0.380 $\pm$ 0.004 & 0.228 $\pm$ 0.004 & 0.353 $\pm$ 0.002 & 0.943 $\pm$ 0.001 & 0.911 $\pm$ 0.003 & 0.918 $\pm$ 0.002\\
		$\tilde{h}_n$(VRNN) & 0.406 $\pm$ 0.003 & 0.261 $\pm$ 0.003 & 0.381 $\pm$ 0.003 & 0.953 $\pm$ 0.000 & 0.928 $\pm$ 0.001 & 0.930 $\pm$ 0.001\\
		$\tilde{h}_n$(VRNN+NN) & \textbf{0.427 $\pm$ 0.003} & \textbf{0.297 $\pm$ 0.004} & \textbf{0.410 $\pm$ 0.003} & \textbf{0.958 $\pm$ 0.001} & \textbf{0.936 $\pm$ 0.001} & \textbf{0.937 $\pm$ 0.000}\\
		\hline
		\multicolumn{7}{c}{Performance with different missing value imputation methods} \\
		\hline
		RNN+NN(zero) & 0.395 $\pm$ 0.005 & 0.248 $\pm$ 0.003 & 0.374 $\pm$ 0.002 & 0.945 $\pm$ 0.003 & 0.918 $\pm$ 0.004 & 0.923 $\pm$ 0.003\\
		RNN+NN(last\&next) & 0.385 $\pm$ 0.002 & 0.233 $\pm$ 0.003 & 0.360 $\pm$ 0.002 & 0.941 $\pm$ 0.001 & 0.912 $\pm$ 0.002 & 0.918 $\pm$ 0.001\\
		RNN+NN(row mean) & 0.393 $\pm$ 0.003 & 0.243 $\pm$ 0.005 & 0.369 $\pm$ 0.001 & 0.945 $\pm$ 0.002 & 0.917 $\pm$ 0.003 & 0.923 $\pm$ 0.002\\
		RNN+NN(NOCB) & 0.384 $\pm$ 0.003 & 0.231 $\pm$ 0.002 & 0.359 $\pm$ 0.001 & 0.941 $\pm$ 0.002 & 0.911 $\pm$ 0.003 & 0.917 $\pm$ 0.002\\
		VRNN+NN & \textbf{0.426 $\pm$ 0.002} & \textbf{0.291 $\pm$ 0.006} & \textbf{0.407 $\pm$ 0.002} & \textbf{0.958 $\pm$ 0.000} & \textbf{0.937 $\pm$ 0.001} & \textbf{0.938 $\pm$ 0.001}\\
		\hline
	\end{tabular}
	\caption{Three sets of diagnosis performances. The top set contains the performances of our models and baselines; the middle set shows the performances of a NN model with features derived from different models; the bottom set compares performance of VRNN+NN and RNN+NN with typical heuristic imputation methods. All results are presented in the "mean value $\pm$ standard deviation" fashion.}
	\label{tab1}
\end{table}

\begin{table}[t]
	\centering
	\begin{tabular}{ccccccc}
		\multicolumn{7}{c}{Performance comparison} \\
		\hline
		Comparison & Micro-F1 & Macro-F1 & Macro-F1-w & Micro-AUC & Macro-AUC & Macro-AUC-w\\
		\hline
		VAE+NN vs. NN &  & * &  & * & * & **\\
		VAE+NN vs. AE+NN &  &  & * & * & * & *\\
		RNN+NN vs. VAE+NN & ** & ** & ** &  & ** & **\\
		VRNN+NN vs. RNN+NN & *** & *** & *** & *** & *** & ***\\
		E($z_n$)(VAE+NN) vs. E($z_n$)(VAE) & ** & *** & *** & ** & *** & ***\\
		$\tilde{h}_n$(VRNN+NN) vs. $\tilde{h}_n$(VRNN) & ** & *** & *** & *** & *** & ***\\
		\hline
	\end{tabular}
	\caption{P-values of paired t-test on diagnosis performances. Note that: *** ($p < 0.001$), ** ($p < 0.01$), * ($p < 0.05$).}
	\label{tab2}
\end{table}

\begin{table}[t]
	\centering
	\begin{tabular}{cc|cc}
		\multicolumn{2}{c}{Different imputation methods} & \multicolumn{2}{c}{Performance comparison}\\
		\hline
		Imputation Methods & Imputation Error & Comparison & P-values  \\
		\hline
		Zero & 0.909 $\pm$ 0.112 & VRNN+NN vs. Zero & *** \\
		Last\&next & 0.434 $\pm$ 0.110 & VRNN+NN vs. Last\&next  & * \\
		Row mean & 0.541 $\pm$ 0.114 & VRNN+NN vs. Row mean & ***\\
		NOCB & 0.547 $\pm$ 0.112 & VRNN+NN vs. NOCB & ***\\
		VRNN+NN & \textbf{0.370 $\pm$ 0.110} & &\\
		\hline
	\end{tabular}
	\caption{The left part is imputation error of different imputation methods. The right part is the performance comparison by paired t-test. Note that: *** ($p < 0.001$), ** ($p < 0.01$), * ($p < 0.05$).}
	\label{tab3}
\end{table}

\section*{Discussion}
From the diagnosis performances shown in the top set of Table~\ref{tab1}, it can be observed that
considering temporal dependencies, e.g., by RNN or VRNN, the diagnosis perform can be significantly improved, confirming that long-term dependency is an important property of clinical data. Secondly, involving deep generative models provides consistent performance improvement (see AE+NN vs. VAE+NN, and RNN+NN vs. VRNN+NN). This is an encouraging result and shows that the generative models (VAE and VRNN) learn better representations compared compared to the less-generative counterparts (AE and RNN).

Compared the improvement provided by the generative modeling, it can be observed that when dealing with single averaged feature vectors (VAE), the improvement is not very significant. This indicates that the data missing problem with the average vectors is not very severe, and the zero imputation is reasonably good. When dealing with temporal vector sequences (VRNN), the improvement is highly significant ($p < 0.001$). This is understandable, as the data missing problem is more serious in this case.

According to the results of the NN classifier with different features (the middle set of Table~\ref{tab1}), we can find that the representations learned from VAE+NN performs significantly better than those learned from VAE; similarly, the representations learned from VRNN+NN are better than those learned from VRNN. This implies that the joint training that considers the classification target can lead to better feature learning. This is not surprising as the learning is now more task-oriented.

The comparison of different missing value imputation methods, as shown in the bottom set of Table~\ref{tab1}, demonstrates that our VRNN+NN model outperforms all the heuristic imputation methods, confirming that the generative model is a more principled way for the missing data treatment. This conclusion can be drawn more explicitly from the results shown in Table~\ref{tab3}, where the performance of different imputation methods is compared by the imputation error directly.
This advantage on data imputation by itself is quite useful in practice. For example, it can help to complete the incomplete laboratory test data and give a rough range of the missing values that may assist physicians to have a better analysis.

\section*{Limitations and Future Works}

\begin{figure}[t]
	\centering
	\includegraphics[width=0.9\linewidth]{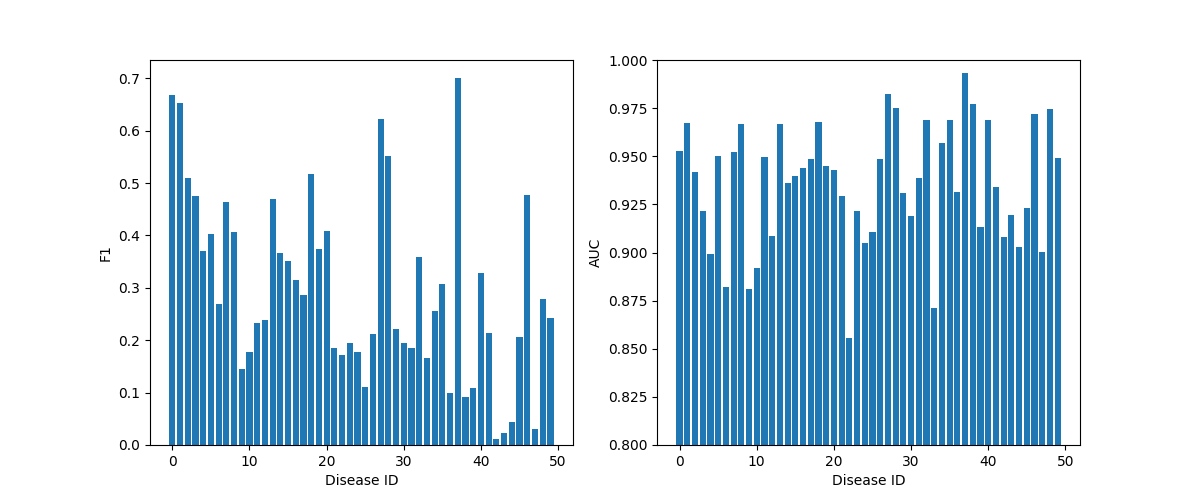}
	\caption{VRNN+NN model diagnosis performance of each diagnosis.}
	\label{fig5}
\end{figure}

Some limitations exist in current study and need to be addressed in future work. First, we only consider the most frequent 50 diagnoses, leaving the rest 2739 diseases untested. This is because there is severe data imbalance between different diseases. About 35\% of 2789 diseases have only one sample. Even for the top 50 diseases selected in the study, the data distribution is still quite biased. As shown in Figure~\ref{fig1}, the disease with ID=0 has 3566 samples, while the disease with ID=49 only has 178 samples. Unbalanced data casts a big challenge for model training, and the resultant models tend to score higher for frequent labels than infrequent ones. In this study, we didn't apply any additional preprocessing, like over-sampling and under-sampling, to address the data unbalance problem. Future work will address the data unbalance problem, by either sampling or re-weighting for infrequent classes.

Second, after looking into the VRNN+NN model performance of each disease, as shown in Figure~\ref{fig5}, we find that the F1 scores and AUCs of different diseases vary dramatically. Some diagnoses, like "Alcohol withdrawal" (ID=37) and "Single liveborn without cesarean section" (ID=0), are quite accurate, while some others, like "Gram-neg septicemia NEC" (ID=42) and "Mal neo upper lobe lung" (ID=47), obtain F1 scores close to 0. This phenomenon implies that some diseases are difficult to detect by our model. It may result from our limitation to 50 most frequent laboratory tests, or the deficiency of other clinical data. More laboratory tests will be considered and other clinical information will be included in the future study.

Third, we only used the in-hospital records that have diagnosis labels, but there are also a lot of outpatient test results in MIMIC-III that posses no labels. In our future work, we will investigate the semi-supervised learning to enhance the capability of our model by utilizing the unlabelled data.

\section*{Conclusion}
The longitudinal, incomplete, and noisy laboratory test data casts a big difficulty for automatically medical diagnosis. In this study, we proposed to utilize a deep sequential generative model in the form of VRNN to deal with the missing data problem and learn the complex temporal patterns in clinical data, and this generative model is trained jointly with the back-end discriminative model for making diagnosis decision. This leads to an end-to-end system that takes advantage of both generative learning and discriminative learning. Our experiments show that the VRNN+NN model significantly ($p<0.001$) surpasses all the baselines and its non-temporal version, the VAE+NN model. We also find that deep generative models can help with imputing missing values in clinical laboratory tests and distilling more informative patterns for diagnosis, and the combination of generative and discriminative models leads to improvement on both feature learning and diagnosis decision. Future work involves addressing data unbalance and utilizing unlabeled data.

\bibliography{sample}
\bibliographystyle{plainnat}


\section*{Author contributions statement}
S.Z. and P.X. conceived and designed this study. S.Z. processed the data and performed the experiments. S.H., P.X., D.W. wrote the paper. D.W. and E.P.X. take responsibility for the paper as co-senior authors. All authors reviewed the manuscript. 

\section*{Additional information}
\textbf{Supplementary information} accompanies with this paper.

\noindent \textbf{Competing financial interests}: The authors declare no competing financial interests.

\end{document}